\documentclass{article}

\usepackage{arxiv}

\usepackage[utf8]{inputenc} % allow utf-8 input
\usepackage[T1]{fontenc}    % use 8-bit T1 fonts
\usepackage{hyperref}       % hyperlinks
\usepackage{url}            % simple URL typesetting
\usepackage{booktabs}       % professional-quality tables
\usepackage{amsfonts}       % blackboard math symbols
\usepackage{nicefrac}       % compact symbols for 1/2, etc.
\usepackage{microtype}      % microtypography
\usepackage{lipsum}
\usepackage{graphicx} \graphicspath{{img/}}
\usepackage[nocompress]{cite}
\usepackage{amsmath,amsfonts}
\usepackage{microtype}
\usepackage{booktabs}
\usepackage{balance}
\usepackage{longtable}
\usepackage{xcolor}

\title{covid-transformer: detecting covid-19 trending topics on twitter using universal sentence encoder}

\author{
  Meysam Asgari-Chenaghlu%\thanks{Use footnote for providing further information about author (webpage, alternative address)---\emph{not} for acknowledging funding agencies.} 
  \\
  Department of Computer Engineering\\
  Tabriz University\\
  \And
  Narjes Nikzad-Khasmakhi\\
  Department of Computer Engineering\\
  Tabriz University\\
  \And
  Shervin Minaee\\
  Machine Learning R\&D\\
  Snap Inc\\
}

\begin{document}
\maketitle

\begin{abstract}
The novel corona-virus disease (also known as COVID-19) has led to a pandemic, impacting more than 200 countries across the globe.
With its global impact, COVID-19 has become a major concern of people almost everywhere, and therefore there are a large number of tweets coming out from every corner of the world, about COVID-19 related topics.
In this work, we try to analyze  the tweets and detect the trending topics and major concerns of people on Twitter, which can enable us to better understand the situation, and devise better planning.
More specifically we propose a model based on the universal sentence encoder to detect the main topics of Tweets in recent months. 
We used universal sentence encoder in order to derive the semantic representation and the similarity of tweets. 
We then used the sentence similarity and their embeddings, and feed them to K-means clustering algorithm to group similar tweets (in semantic sense).
After that, the cluster summary is obtained using a text summarization algorithm based on deep learning, which can uncover the underlying topics of each cluster.
Through experimental results, we show that our model can detect very informative topics, by processing a large number of tweets on sentence level (which can preserve the overall meaning of the tweets).
Since this framework has no restriction on specific data distribution, it can be used to detect trending topics from any other social media and any other context rather than COVID-19. Experimental results show superiority of our proposed approach to other baselines, including TF-IDF, and latent Dirichlet allocation (LDA).
\end{abstract}

% keywords can be removed

\section{Introduction}
\begin{figure}[h]
\begin{center}
   \includegraphics[page=2,width=0.8\linewidth]{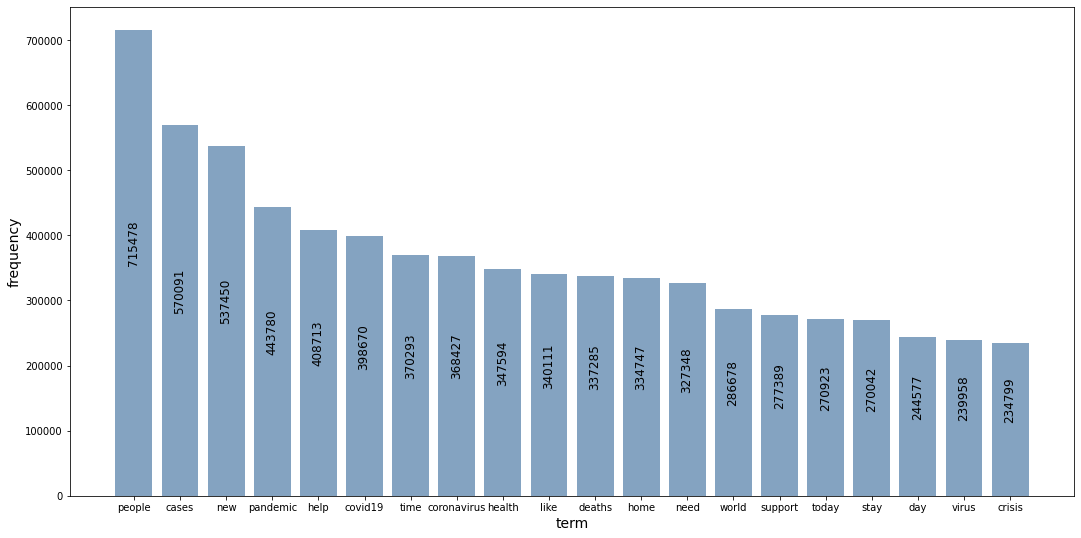}
\end{center}
   \caption{The most frequent words used by people on Twitter during the first 3 months of pandemic.}
\label{frequent_words}
\end{figure}

COVID-19 has led to a global pandemic, impacting more than 200 countries across the globe, infecting more than 20 million people, and causing more than 750,000 death as of Aug 12, 2020 \cite{worldmeter}, and there is a large number of research works 
With the growing scale of COVID-19 (after March 2020), there has been a shift in the distributions of Tweets posted on Twitter, reflecting the fact that COVID-19 has become a major concern of people across the world.
To illustrate this, in Figure \ref{frequent_words} we show the most frequent words used in Twitter %\textcolor{blue}{(shervin: I guess the dataset you used has already kept covid-19 related tweets only, so these are not the only popular words.)}\textcolor{red}{Meysam: The dataset we have used has only 33 days of the tweets related to covid.} 
during April 2020 (the 2nd month of global pandemic of COVID-19). As we can see words like "death", "help", "COVID-19" are among the popular words.

Analyzing people's opinion and concerns on social media can help us to better understand public's concern and expectations, enabling the government and health official to have a better planning for managing the situation.
There has already been a few works analyzing COVID-19 related Tweets. 
As an example, in \cite{kleinberg2020measuring}, Kleinberg et al. presented a ground truth dataset of emotional responses to
COVID-19, and presented a framework to detect the main concerns of people in different countries on COVID-19 subject.
In \cite{ordun2020exploratory}, Ordun and colleagues analyzed five different techniques to assess the distinctiveness of topics, key terms and features, speed of information dissemination, and network behaviors for Covid-19
related tweets.
In \cite{singh2020first}, Singh et al. looked at the information and misinformation shared on Twitter. 
They tried to see if the discussion is emerging from, myths shared about the virus, and how much of it is connected to other high and low quality information on the Internet.
So far, there has not been a solid framework which leverages the state-of-the-arts in natural language processing, to detect the trending COVID-19 related topics in social medial.
Besides works trying to analyze textual data about COVID-19, there are many works trying to analyze other types of data (images, time-series, clinical information) to get insight and predictive models about different aspects of COVID-19 \cite{covid_res1, covid_res2, covid_res3, covid_res4}.

In this work, we propose a deep learning based framework to detect trending topics people are talking about on Twitter, using a combination of transfer learning and clustering algorithms.
Deep learning based models have been very successful in achieving state-of-the-art results in many of the NLP problems in the recent years, including word embedding, sentiment analysis, question answering, and machine translation \cite{young2018recent, bahdanau2014neural, minaee2017automatic}.
We first extract the representation (embedding) of sentences in Tweets using "Sentence Transformer", which can capture the semantic information of the sentences. 
We then use clustering algorithms to group similar sentences (based on their embeddings) into the same groups. Ideally different clusters contain different semantic topics. 
After that, text summarization is used to get the summary of sentences in each cluster, which can uncover the trending topics of them.
%After that, the most representative sentences and words of each cluster are extracted, to find the trending topics.
%Two approaches are used to derive the representative words of each cluster, one based on word frequency, and the other one based one the closeness of that word to the other words.

\begin{figure}[h]
\begin{center}
   \includegraphics[page=2,width=0.66\linewidth]{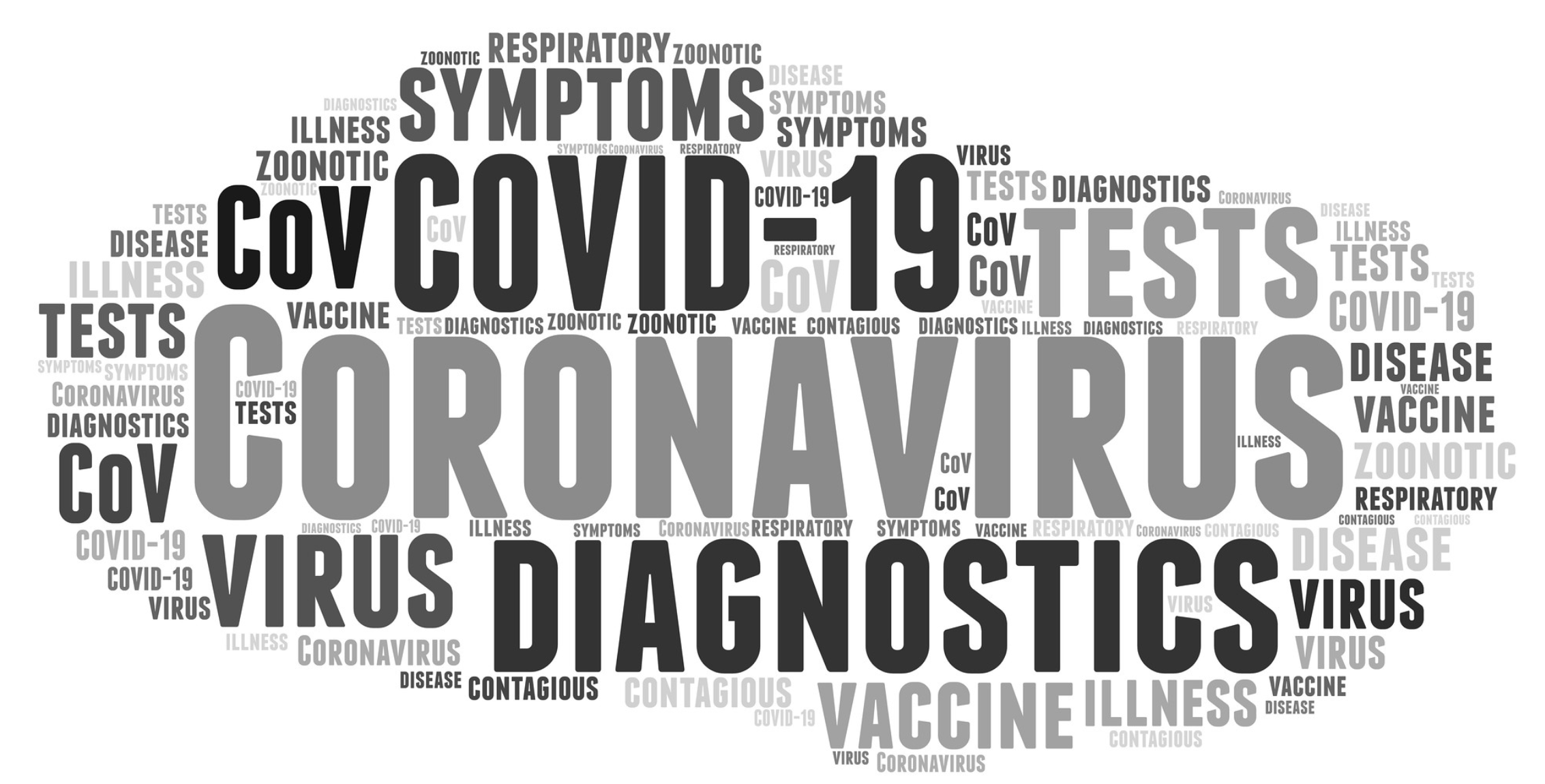}
\end{center}
   \caption{The wordcloud of popular words on Twitter during the first few months of pandemic. Courtesy of UPenn.
   %https://penntoday.upenn.edu/news/coronavirus-or-covid-glossary-help-navigate-pandemic-vocabulary
   %\textcolor{blue}{shervin: can you please update this figure with word-cloud derived from our entire dataset (perhaps subsampled one)? This one is taken from internet!}
   }
\label{clus1_wordcloud}
\end{figure}

Compared to the classical models for topic modeling (such latent Dirichlet allocation, LDA), this work better employs the semantic information and meaning of the tweets, by first representing a sentence-level embedding of tweets, and then using those embeddings to group the tweets. 
In this way, we can find similar topics by directly analyzing sentences, rather than considering words similarity (as used in LDA).
Figure \ref{clus1_wordcloud} shows the word-cloud of sentences belonging to one of the topic clusters of our model.

Here are the main contributions of this work:
\begin{itemize}
    \item We provide a novel framework, which can detect the trending COVID-19 related topics on social media, such as Twitter, in an unsupervised fashion. We do so, by first extracting a sentence-level embedding using sentence-Transformer, and then grouping similar sentences into the same cluster using k-means clustering, and finally extracting each cluster summary with a deep learning based text summarization. This summary contains the major topics of each cluster.
    \item We provide a detailed experimental study showing the promise of this work, and its advantages over simple baselines such as using TF-IDF and latent Dirichlet allocation (LDA). 
\end{itemize}

The structure of the rest of this paper is as follows: In section \ref{sec:method}, we provide the details of the proposed algorithm. 
Section \ref{dataset} provides an overview of the dataset used in our experiments. 
In Section \ref{sec:experimental_setup}, we provide the experimental analysis of the proposed algorithm in terms of detected topics, trending words in each topic cluster, and sentence representation similarity. 
And finally the paper is concluded in Section \ref{conclusion}.

\section{Proposed method}
\label{sec:method}
\begin{figure}[h]
\begin{center}
   \includegraphics[width=0.8\linewidth]{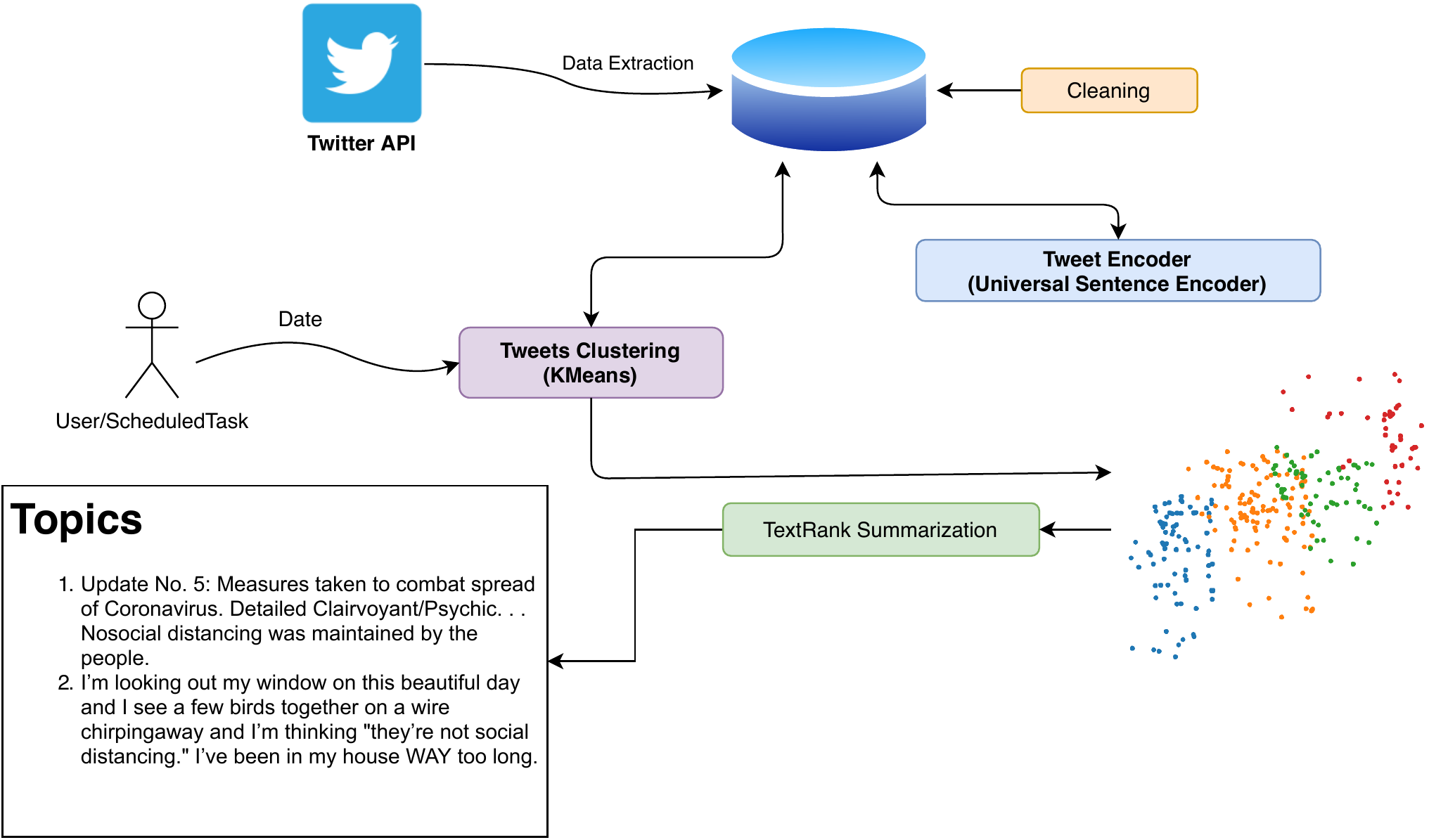}
\end{center}
   \caption{The overall structure of our proposed method.}
   \label{fig:overall_str}
\end{figure}

This work proposes a new framework for detecting the trending COVID-19 related topics on Twitter, using Universal Sentence Encoder and text summarization.
The overall structure of our proposed approach is presented in Figure \ref{fig:overall_str}.
As it can be seen from this figure, we first obtain the adequate data from Twitter and get tweets through twitter API. 
After that, data cleaning part needs to performed on the collected tweets. 
Then, the Universal Sentence Encoder is used to extract the feature representation (embedding) of tweet's sentences. 
%Then, the Universal Sentence Encoder is used to extract the feature representation (embedding) of tweet's sentences. 
%Then, the Universal Sentence Encoder is used to extract the feature representation (embedding) of tweet's sentences. 
The sentence embeddings from different tweets are then fed into the k-means clustering algorithm to group them semantically similar ones into the same cluster.
%This semantic embedding provides a semantic mapping from tweets. 
%These embeddings are stored on the database. 
%Afterward, a user defines a period of time and the embeddings of tweets that exist in the time span are retrieved from database. 
%The next step is detecting the trending topics of COVID-19 and semantically discriminating between different tweets. 
%This discrimination can be approached by any method such as KMeans. 
In the end, TextRank summarization technique is applied on the sentences of cluster to generate a summary of each cluster, which contains the most representative topic.  
More detailed description of each stem, is provided in the following subsections.

\subsection{Tweet Encoder}
In recent years, there have been many efforts  to create a semantic representation of textual sequence (such as sentence, paragraph, or document). 
These methods include a wide range of techniques from word movers distance to the recent state-of-the-art methods such as SentBert and Universal Sentence Encoder \cite{kusner2015word,reimers2019sentence,cer2018universal, minaee2020deep,asgari2020topicbert}. 
All of these methods aim to provide a vector representation of sentence which can capture its semantic meaning, and provides similar representations for similar sentences.

\begin{figure}[h]
\begin{center}
   \includegraphics[width=0.8\linewidth]{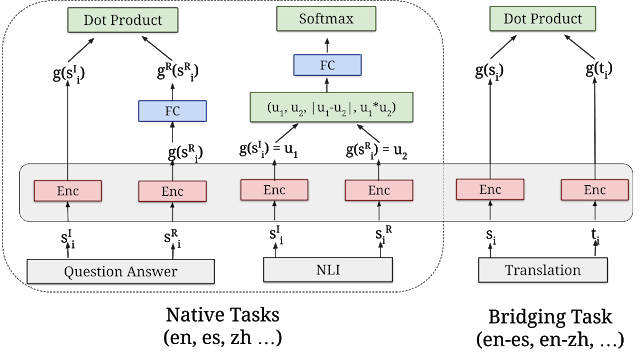}
\end{center}
   \caption{Architecture of \text{USE\_T} and its multi-task/multi-lingual learning paradigm with shared parameters \cite{cer2018universal}.}
   \label{fig:transformer}
\end{figure}

For example, Sentence BERT (also known as SentBERT), uses a Siamese coupled neural network composed of two identical instances of BERT. Like any Siamese neural network, this model aims to find the similarity between the two inputs sentences. 
Universal sentence encoder is another text embedding technique that has different versions for different use cases. 
One variation of this model was trained with Deep Averaging Network (DAN), called \texttt{USE\_DAN}. This encoding model produces sentence embeddings in the way that it first gets the average embeddings of words and bi-grams, and then applies a feed-forward neural network on the average representation. 
On the other hand, newer version of universal sentence encoder is based on the Transformers%\textcolor{red}{I corrected the sentence} \textcolor{blue}{shervin: the explanation of this sentences is very vague, is this a new model, or a variation of the previous one? there should be more clear transition from one sentence to another!}\textcolor{red}{Meysam: Corrected by adding new version phrase to indicate that it had an older and newer versions}
, called \texttt{USE\_T}, that has higher accuracy and is computationally more intensive than \texttt{USE\_DAN}. In this research, we employ the transformer encoder version of universal sentence encoder. \texttt{USE\_T} can handle words, sentence and documents, as the input. 
Figure \ref{fig:transformer} shows the architecture of this model and how it encodes text into high dimensional vectors. As it can be seen from the figure, \texttt{USE\_T} has been trained on various downstream tasks. The encoder blocks in this figure are based on the Transformer model proposed by the Vaswani et. al\cite{vaswani2017attention}.

The multi-task and multi-lingual training paradigm of the \texttt{USE\_T} makes it more suitable for tasks such as semantic sentence pair retrieval. 
On the other hand, deploying \texttt{USE\_T} in our study is more crucial with respect to relation of semantically close tweets. 
The most powerful side of this architecture is its ability to find similar texts without need for any pair computations. 
%Instead of using a Siamese model that gets two distinct texts (size or length of the text is not issue of discussion here) and computes the cosine similarity or any other distance, this method represents each text in a dense vector format. 
It provides a dense vector representation of each text unit, and this dense vectors can be used for computing distance or similarity between different tweets or sentences \cite{cer-etal-2018-universal,yang2019multilingual}. 
%In other words, this architecture, transforms any given text to a dense vector space that it is easier to grasp the semantic similarity compared to other methods.

Based on the above reasons, we employ universal sentence encoder to capture the sentence representation of tweets. The tweet embeddings obtained from \texttt{USE\_T} enables us to calculate the similarity of sentences in a semantic way, which is used by the clustering phase. It should be noted that before injecting tweets as the input of \texttt{USE\_T}, text cleaning is a must-do process and  we performed this by selecting the most sensible and usable sentences from the pile of tweets. 
The sentences which contain hashtags and mentions, are removed from our corpus.

\subsection{Tweets Clustering}
As it can be seen from Fig \ref{fig:overall_str}, once the embedding are extracted, a clustering algorithm is used to cluster tweet's sentences based on their embeddings obtained from previous step. 
The distance between two sentences's embeddings in this case, provides the dissimilarity of tweets. 

Different clustering algorithms can be used for this purpose, such as K-means, spectral clustering, mean-shift, Density-Based Spatial Clustering (DBSC). 
We use K-means clustering algorithm here, for its simplicity, speed, and the ability to pre-define the number of clusters.

\subsection{Cluster summarization}\label{cluster_summarization}
The clustering step, provides several groups of semantically similar sentences.
On high level, the tweets in the same groups should contain more similar topics, than those in different groups. 
Although the centroid of each cluster should contain the average embedding (therefore the average topic/concept of that cluster), it will not necessarily capture all topics of that cluster. But it could serve as a simple baseline.
%A  naive but effective method to show the topic of each cluster is selecting the most close data point to the cluster centroid that can be named as the repsentative tweet. By doing such naive method, we would lose the most information in other tweets that is the words. 
A better solution for finding the topic of each cluster is to use a text summarization technique to provide a meaningful and sensible summary of that cluster, capturing the key topics. 
Here we used the TextRank summarization framework.
More details on TextRank summarization is provided in  \cite{barrios2016variations,mihalcea-tarau-2004-textrank}. 
%The details of The results present sentences for each topic that best describes the topic based on that specific cluster tweets.

\section{Dataset}
\label{dataset}
To evaluate the performance of the proposed framework, we used a dataset of tweets between 2020-03-29 and 2020-04-30, that are collected via twitter API. 
This dataset includes more than 8 million tweets in total for English language. 
We only used a random sampled of 20 percent of this dataset, which contains 1.6M tweets in total.

\begin{figure}[h]
\begin{center}
   \includegraphics[width=0.8\linewidth]{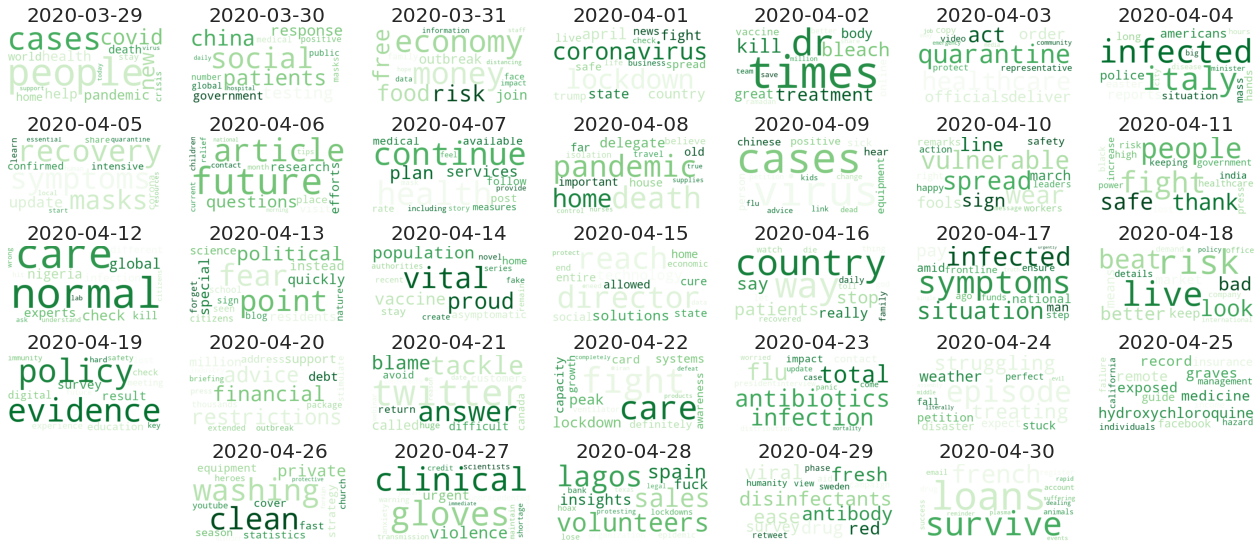}
\end{center}
   \caption{Wordcloud of twitter corona dataset based on each day; Each row presents a week. %\textcolor{blue}{shervin: I think it may e better to put the word "people" in stopwords, as right now this is the biggest word of most of the clouds. Ideally we want to have medical related words, financial related words, political related words, etc. in different clusters/wordclouds. Also please shift the last row with one figure, so we get a semantic visual look.} 
   }
   \label{weekly_wc}
\end{figure}

%Figures \ref{weekly_wc}, \ref{daily_freq} and \ref{boxplot} provide some aggregate information about this  dataset. 
Figure  \ref{weekly_wc} denotes the wordcloud of the most frequent words after removing the stopwords for each day. 
One interesting fact from this plot is that, some words are always present among the popular ones for different days. We removed any repeatative words of previous days to have  visualization in this figure.

\begin{figure}[h]
\begin{center}
     %a.\\
     %\includegraphics[width=0.5\linewidth]{img/tweet_frequency.pdf} \\
    %b.
    %\\
    \includegraphics[width=0.8\linewidth]{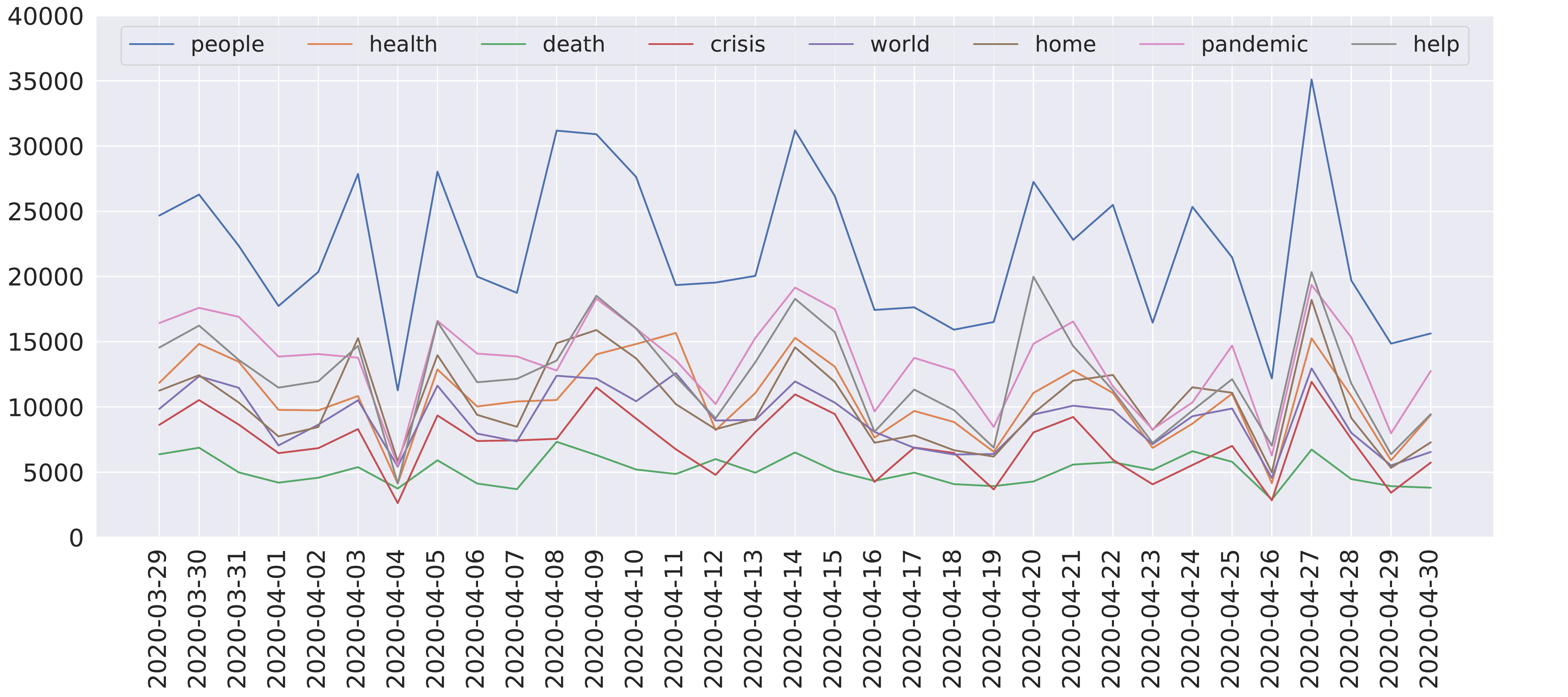}
    
\end{center}
   \caption{
   %a. Trending Topics chart throughout 33 days; b. 
   Daily frequency of top frequent words in Tweets. %\textcolor{blue}{shervin: I suggest to show the frequency of 5-8 popular words over different days here. So we can see how the temporal trend of those words have changed, rather than overall trend.}
   }
   \label{daily_freq}
\end{figure}

\begin{figure}[h]
\begin{center}
   \includegraphics[width=0.9\linewidth]{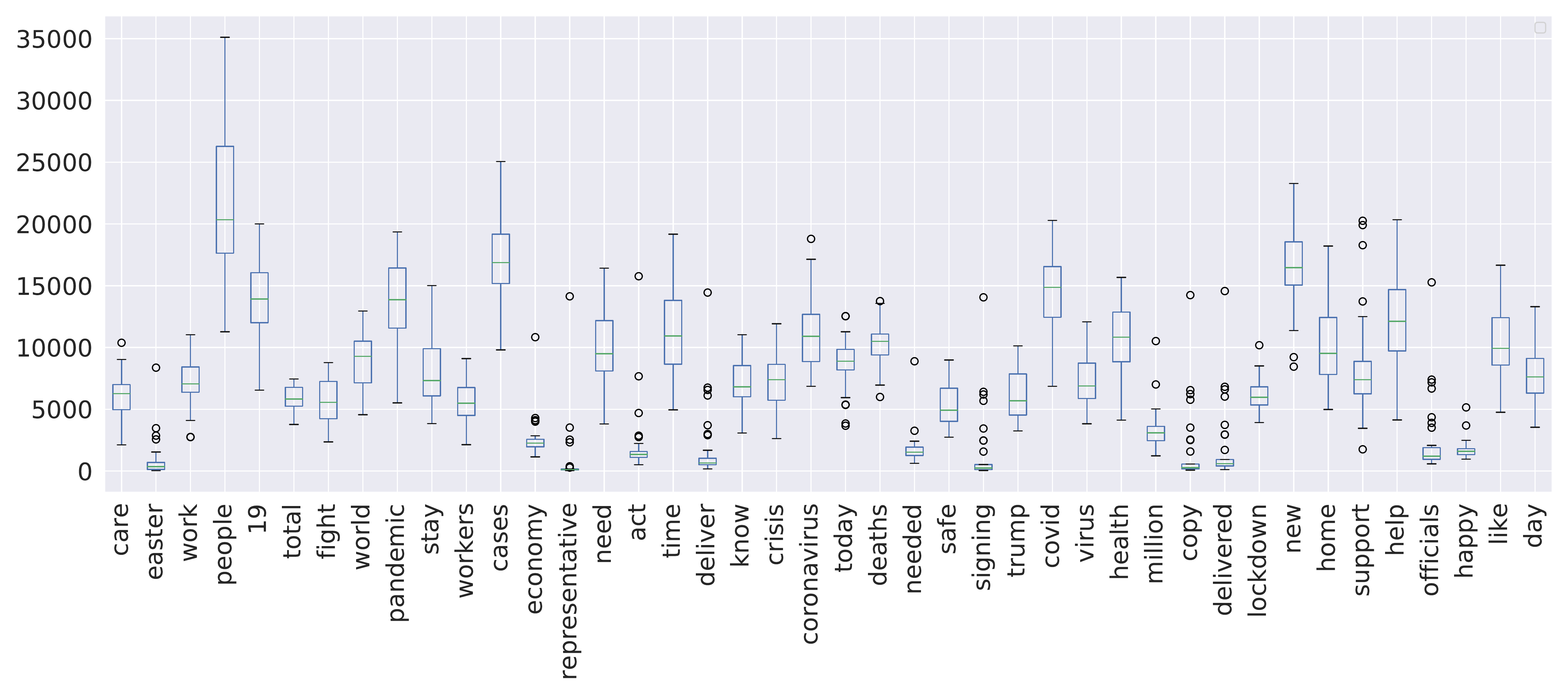}
\end{center}
   \caption{Box plot of word frequencies over time; The most frequent words in each day are selected.}
   \label{boxplot}
\end{figure}

%Also, Figure \ref{daily_freq} (a) shows the frequency of tweets for each of these 33 days. Dramatic raises in tweet frequency and overall decrease in frequency of tweets demonstrate the up and downs of the user posts in this topic. As it can be seen from the figure the first few days of covid this topic has attracted more attention, while there is a considerable decrease in the middle of time span. After that, covid topic has become a hot topic once more. Moreover, 
Figure \ref{daily_freq} shows the frequency of 8 popular words over different days. So we can see how the temporal trend of those words changes  over time. 
A box-plot visulization is also presented in Figure \ref{boxplot}, which provides information about the most frequent words for the entire dataset. 
Th distribution shown in this plot is acquire over 33 days. 
For some words such as \texttt{singing}, only the first days has dramatic frequency raise, while for other days, this frequency drops seemingly. Specific words such as \texttt{help} are present for entire dataset and distribution is close to normal.

We also labeled the dataset using opinions of six experts. The labels have been collected as group of words for each topic and major news media headlines have been used.

\section{Experimental Study}
\label{sec:experimental_setup}
In this section we provide the experimental study details, and the model results.
Before going into the model analysis, we will provide the details of the hyper-parameter values, and experimental setup.

\subsection{Experimental Setup}
We used the universal sentence encoder transformer based version 5, which is available at Google Tensorflow-hub \footnote{\url{https://tfhub.dev/google/universal-sentence-encoder-large/5}}. 
This version has been trained on a monolingual dataset with embeddding dimension of 512. We used dot product for similarity of embeddings.

For K-means clustering algorithm, the number of clusters is set to 30, and the max iteration to 300.
This algorithms has been initialized with K-Means++ algorithm.  
After clustering, we merged smaller clusters with the larger ones using the similarity of their centroids. 
The clusters with smaller sizes are thought to be noisy ones. 
The clustering algorithm has been applied for each day (out of the 33 days) and the summarization is performed on the largest clusters in each day. 
The maximum summarization word count is set to 20 words and the TextRank based method is used for this task. 
The resulted summary for each day contains two or three sentences.
%\textcolor{blue}{shervin: I wonder if this is the best way to capture the trending topic of each day? On one hand we only used one cluster, and on the other hand each summary contains 2-3 sentences. How can this capture a case where there are more than 5-6 distinct topics?}\textcolor{red}{Meysam: Instead of just triggering most important words, we proposed two distinct approaches to present the results of clusters: 1) conventional words as topics which everybody does, 2) summarising the clusters which just a few people did to prove the quality of their cluster/topic detection algorithm.}

\subsection{Experimental Results and discussion}
This section presents the experimental results. Before revealing the outputs, we should note some points . 
First point is that the evaluation is performed on the subset of 20\% of the samples. The second point is that we use TF-IDF for baseline comparison. 
Finally, the same clustering method (K-means) and summarization technique (using same TextRank as  described in sec. \ref{cluster_summarization}) are applied to TF-IDF approach. 
For LDA its default setup is used for comparison.

\subsection{Model Results}
\label{results}
\begin{figure}[h]
\begin{center}
   \includegraphics[width=0.9\linewidth]{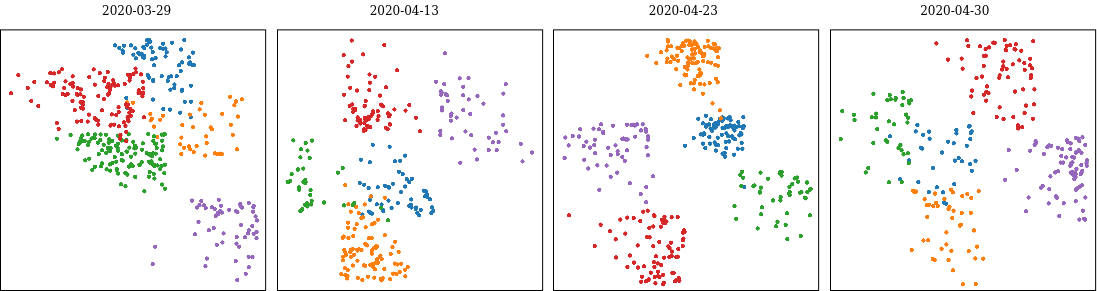}
\end{center}
   \caption{UMAP visualization of topics for each day; we dropped the outliers of each cluster for better visualization.%\textcolor{blue}{shervin: I suggest to show this T-SNE for 3-4 days only, but make it bigger. And for those 3-4 days, please change the T-SNE parameter so that we can see a more clear separation between different clusters. If we can show a few words from each cluster on top of it, it would be even better, but I leave it optional. If T-SNE does not given you good results, try  UMAP instead.}\textcolor{red}{Meysam: I assume this is satisfying :D}
   }
   \label{fig:results_clusters}
\end{figure}

\begin{figure}[h]
\begin{center}
   a.\includegraphics[width=0.4\linewidth]{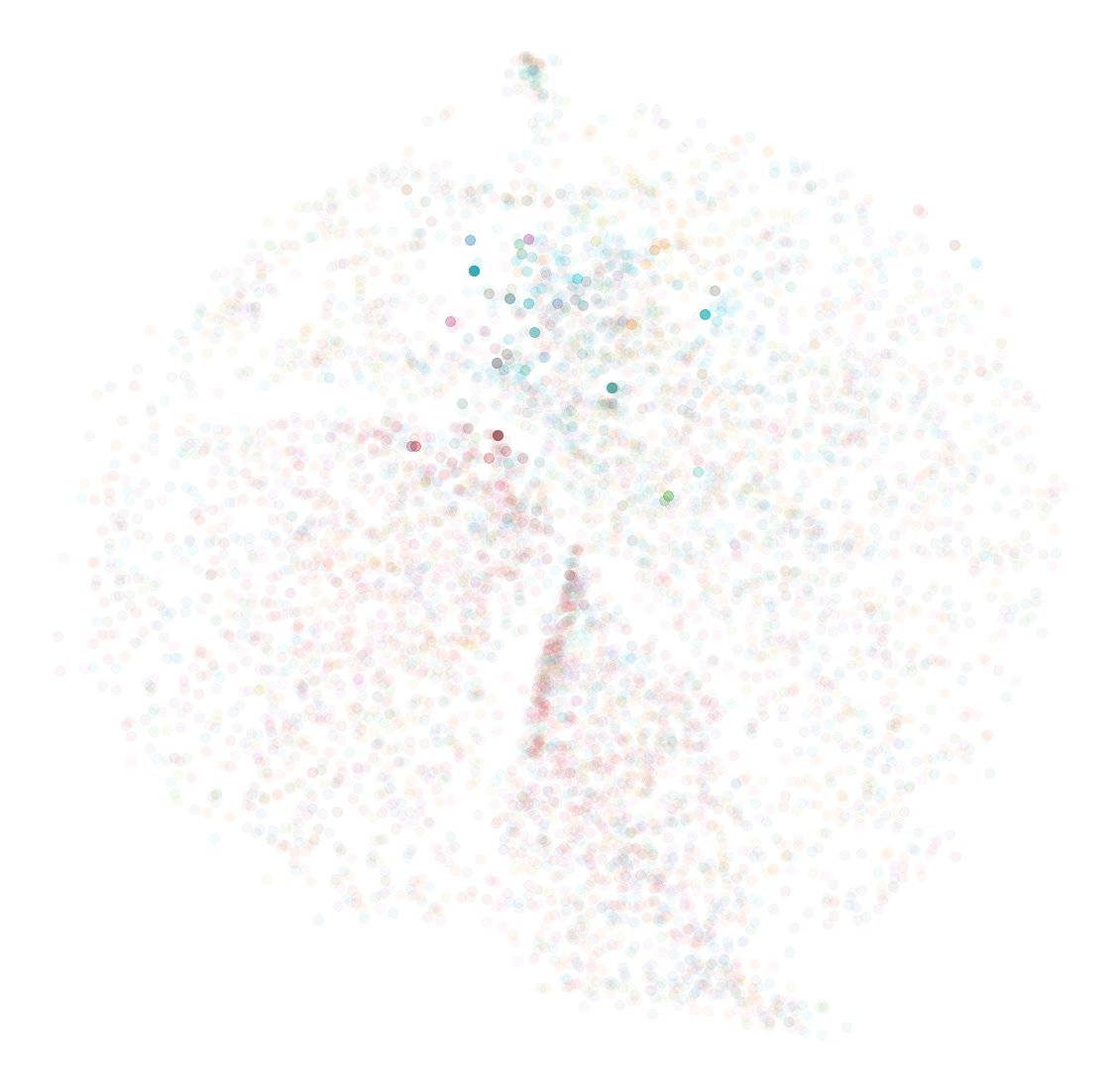}
   b.\includegraphics[width=0.4\linewidth]{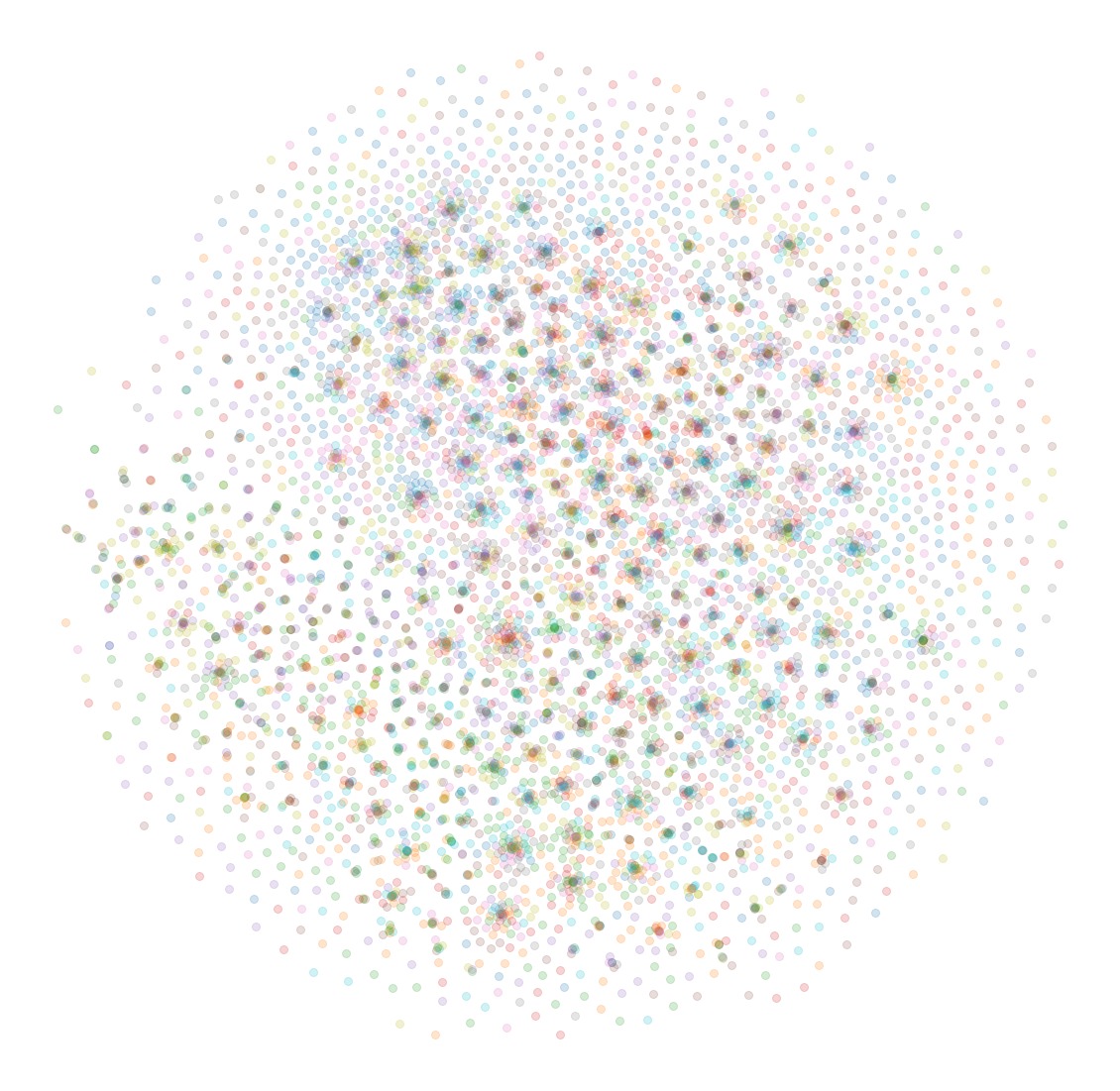}
\end{center}
   \caption{Cluster results using UMAP dimensionality reduction for the first day of dataset: a. TFIDF; b. Our Approach  
   %\textcolor{blue}{shervin: can we also show T-SNE of our work side-by-side to see the difference?}
   }
   \label{fig:tfidf_baseline}
\end{figure}

Figure \ref{fig:results_clusters} denotes the UMAP visualization of the tweets embedding of the most representative cluster for four days in our dataset.
This figure shows five major clusters for each day. The presented UMAP dimensionality reduction shows the separability of clusters. Considering overlaps of all clusters on a major topic (covid19), this topic separation is an advantage. 
%\textcolor{blue}{shervin: more discussion on the main observations from this figure is needed!!!}
Figures \ref{fig:tfidf_baseline} (a) and (b) illustrate the visualization results of TF-IDF and our approach for the first day of dataset. As we can see our approach provides more distinctive groups in this cluster.
%Moreover, the visualization part is more sensible. 
For the sake of comparison and producing a baseline for this dataset, we keep the original pipeline and use TF-IDF instead of universal sentence encoder. 
As it can be derived from the Figure \ref{fig:tfidf_baseline} (a), the clusters created by TF-IDF are more separable and it is because of TF-IDF nature. But in case of output results, it did not have better result compared to our approach because of its inability to capture semantic relevance of inter-topic a irrelevance of out of topic texts. 
In another word,  TF-IDF completely separates topics that we know is not correct due to the fact that there are significant overlaps among topics %\textcolor{blue}{shervin: you should support your claim with some experimental results! where can we see that TF-IDF generates worse topics than us?}\textcolor{green}{Meysam: We did compare it in table \ref{tb:q_results}}. 
Figure \ref{fig:tfidf_baseline} (b) proves that our approach separates topics by considering their overlapping. That means our approach obtains overlapping communities in which the tweets have the same topics.
 
Summarized text obtained by our method for each topic for all days is presented in Table \ref{tab:summarized_topics_full}. 
Table \ref{tab:tfidf_topics} demonstrates the resulting topics and extracted keywords using TF-IDF, which can be compared to Table \ref{tab:summarized_topics_summarized} for the first date of dataset incident. 
These tables prove that using universal sentence encoder not only enables a better clustering result, but also makes a better summary of each cluster than TF-IDF approach. The obtained keywords also are another evidence to confirm \texttt{USE\_T} works better than TF-IDF.  \texttt{USE\_T} suggests meaningful keywords which discover main topics in tweets. 
While, TF-IDF can't capture semantics and detect the most relevant keywords that represent the topic of the cluster.

\begin{table}[]
    \centering
    \begin{tabular}{|c|p{8cm}|l|}
    \hline
     Date/Topic No.& Topic Summary & Topic Keywords  \\\hline
     2020-03-29/1&US desperately needs to start doing this to help prioritize where to send new vents Farm across the road, all day Friday, yesterday, \& today, sound tractors.&day; US; help;\\\hline
     2020-03-29/2&
*Highly contagious, deadly virus in the UK* Some idiot: "Let's send every house in the country a letter." *Touches millions of letters* Last night, together with local supporters, we distributed cooked food to 200 families in Ahmedabad, Gujarat. & contagious; virus; house;\\\hline
2020-03-29/3&I know lots of people are saying and feeling this but you can never say thank you enough- thank you to all the essential workers, especially health care, pharmacies \& grocers. & people; pharmacy; health;\\\hline
\end{tabular}
    \caption{TFIDF results of proposed pipeline.}
    \label{tab:tfidf_topics}
\end{table}

\begin{table}[]
    \centering
    \begin{tabular}{|c|p{8cm}|l|}
    \hline
     Date/Topic No.& Topic Summary & Topic Keywords \\\hline
     2020-03-29/1&Here are yesterday’s state by state numbers: This might be relevant information for a certain world leader in a country called the “United” states of America.&numbers; US; leader; \\ \hline
2020-03-29/2&People may get infected, possibly even die because of selfish, stupid idiots like this.& infected; die; selfish;\\ \hline
\end{tabular}
    \caption{Results of our proposed approach.}
    \label{tab:summarized_topics_summarized}
\end{table}

Table \ref{tb:q_results} shows the quantitative results of our approach compared to LDA and TF-IDF. We used the labeled version of the dataset for this comparison. Top 10 precision, recall and f1 measure are used as metrics. From this table, it is clearly seen that our approach made over 11 and 27 percent improvements compared to LDA and TF-IDF respectively.
%We also compute the 95\% confidence interval of these rates, to see if the obtained gains are statistically significant. 

\begin{table}[]
    \centering
    \begin{tabular}{|c|l|l|l|}
    \hline
     Method& P@10 & Re@10 & F1@10 \\\hline
     
     Our Approach&\textbf{0.67} %$\pm$ 0.004}
     &\textbf{0.58}% $\pm$ 0.005}
     &\textbf{0.62}% $$\pm$ 0.005}
     \\ 
     \hline
     
    TFIDF & 0.31 %$\pm$ 0.004 
    & 0.41 %$\pm$ 0.005 
    & 0.35 %$\pm$ 0.005
    \\ 
    \hline
    
    LDA & 0.54 % $\pm$ 0.005 
    & 0.49 % $\pm$ 0.005 
    & 0.51 % $\pm$ 0.005
    \\
    \hline
\end{tabular}
    \caption{The quantitative comparison of the performance of the proposed model and two popular baseline for topic extraction. 
    %The confidence interval of these rates are also provided.
    }
    \label{tb:q_results}
\end{table}

\subsection{Discussion}
As it is clear from the visualization outputs of TF-IDF, the clusters tend to be at the very same center or they are some noisy looking points around a centroid. 
In case of our approach, the clusters are divided into more separated sub-groups. The reason is that our approach considers and obtains semantic features by utilizing the \texttt{USE\_T} for generating embeddings while TF-IDF produces raw vectorization. 
On the other hand, the result of TF-IDF in table \ref{tab:tfidf_topics} shows that it is merging topics that are not combinable. 
In contrast, in our approach, the clusters can be combined, because of semantic relativity and the dot product similarity metric, which in the case of TF-IDF no such approach would be applicable. Briefly, the partitions originated by our proposed method is more meaningful than TF-IDF due to extracting more semantic features from text and considering overlaps between topics.

Moreover, the semantically close points from output of \texttt{USE\_T} shows groups of near text representations that are centered around specific centroids with some kind of noise around them. 
While in case of TF-IDF, the noise is all over the reduced dimension output and no visually detectable center points are available (discarding the same points transformed to same location in some points that means they were the very same sentences with very close or same words).

Another deduction from the results is that due to the emphasis of \texttt{USE\_T} on semantic representation of tweets, it produces better quality and meaningful summaries of each topic cluster rather than TF-IDF method, which considers statistics of words of the tweets. Also, our approach finds keywords that better present the most relevant topics contained in the cluster.

Clustering or topic detection based on bag-of-words methods and sparse term frequency vectorization such as TFIDF cannot capture semantic relations. LDA on the other hand, lacks the same capability. Results from table \ref{tb:q_results} obviously shows this gap between our approach and these two methods.

\section{Conclusion}
\label{conclusion}
In this work, we proposed a trending topic detection framework using a method that combines the Transformers with text summarization in a smart way, and applied that to COVID-19 related Tweets on Twitter. 
First, the sentence embeddings are extracted using Transformer. Then these embeddings are fed into a clustering algorithm to group similar Tweets, and finally text summarization is applied to all sentences of a each cluster to provde a short summary of that.
This framework can provide us the resulting topics in the form of concise sentences that are easier to read and comprehend for human. 
This approach is applied to COVID-19 pandemic dataset collected from twitter, and several experimental studies are performed to assess its performance. 
Through experimental comparison, we showed that this model outperforms other popular topic detection approaches, based on TF-IDF, and LDA.

\bibliographystyle{unsrt}  

%%% Comment out this section when you 
%\bibliography{references}

\appendix
\section{Appendix}
\begin{longtable}{|c|p{14cm}|}
\hline
Date/Topic No.& Topic Summary\\
\hline
2020-03-29/1&Here are yesterday’s state by state numbers: This might be relevant information for a certain world leader in a country called the “United” states of America.\\ \hline
2020-03-29/2&People may get infected, possibly even die because of selfish, stupid idiots like this.\\ \hline
2020-03-30/3&People are out there looking for mere yaleero "We're just trying to spread a bit of joy." Six years back, I put my brain to thinking about what the world would really be like at the end.\\ \hline
2020-03-30/4&I'm 5ft 8' \\& altho' the wobble base might yield a more energetic game, it looks like it's shorter than the Pro. Help!\\ \hline
2020-03-31/5&A municipal worker sprays disinfectant to prevent the spread of LIVE UPDATES: Cape May closes city beaches amid spread of COVID-19.\\ \hline
2020-03-31/6&Looks like I'm gonna need a letter showing it's ok for me to be on my way to work.\\ \hline
2020-04-01/7&We don't have enough ventilators, health care workers, personal protective equipment and hospital beds, the governor says.\\ \hline
2020-04-01/8&This government is failing its people!\\ \hline
2020-04-02/9&Freddie, 16, said isolation feels like living in a box Still not reached the first target of 10000/day.\\ \hline
2020-04-02/10&There is no bottom with Trump.
Why didn't Trump?
To Trump, EVERYTHING is about him.\\ \hline
2020-04-03/11&The good things about today is that it’s Friday, it’s my day off, I got food and liquor and Money Heist season 4 is released.\\ \hline
2020-04-03/12&"INFO CORONAVIRUS N-COVID19 CHINAVIRUS WAKE UP WORLD.." Broadband engineers threatened due to 5G coronavirus conspiracies Part three: the church's social mission.\\ \hline
2020-04-03/13&"Not ready for primetime...dont want to give people false hope" A lot of them probably don’t know they’ve even had it Is it just droplets?\\ \hline
2020-04-04/14&From "oh, hopefully no big deal", to "uh, that looks suspicious" slowly creeping to "holy shit, I'm gonna lock myself away from all this" 2020 is definitely going to change what normal life means.\\ \hline
2020-04-04/15&Sopore Adminstration discharged 23 students after completing 14-days of quarantine at Sopore Hospital during  government imposed nationwide lockdown as a preventive measures against the spread of the COVID-19 coronavirus, in Kashmir on April 4,2020.\\ \hline
2020-04-05/16&Update No. 5: Measures taken to combat spread of Coronavirus.
Detailed Clairvoyant/Psychic… No social distancing was maintained by the people.\\ \hline
2020-04-05/17&I'm looking out my window on this beautiful day and I see a few birds together on a wire chirping away and I'm thinking "they're not social distancing."  I've been in my house WAY too long.\\ \hline
2020-04-05/18&Way people expect State to provide for them currently is a communist concept Focusing on moving from disorder in today's world The Cynefin framework significantly helps us to determine what particular parts we are dealing with, in the decisions needed.\\ \hline
2020-04-06/19&“Think of older people as the cavalry coming over the hill.” so I think it also work on corona.\\ \hline
2020-04-06/20&If You’re a true patriot then UNITE all ppl of all nations for the governments of the world have planned to enslave you all Is it because they want to keep Americans sick, and dying, for political reasons?\\ \hline
2020-04-06/21&YOU CANNOT HAVE A VACCINE FOR THIS VIRUS.\\ \hline
2020-04-07/22&After this 24/7 lockdown, the next time we're able to go anywhere near a beach, I better be looking something like this.\\ \hline
2020-04-07/23&. here's another 10 from today Hear from our experts during our live Q\\&A on April 9: The next impeachment trial gonna be Lit. Note the time is AM, Ontario Canada.\\ \hline
2020-04-07/24&Thank you all doctors, nurses and paramedical staffs for stay far away from home and protecting our lives from this pandemic corona virus.\\ \hline
2020-04-08/25&Hard days, weeks, months, years, lives.
Coronavirus world update Today 8 April Stay at home Like...\\ \hline
2020-04-08/26&government spent millions of money to facilitate the huduma number process, why can't government use the data which was collected to disburse funds to cater for kenyas daily needs Whatever govt decides will be in favour of country \\& citizens.\\ \hline
2020-04-09/27&Q He's what a real leader is like.
Coronavirus (COVID-19): How Does It Impact Commercial Leases?\\ \hline
2020-04-09/28&Yeah, little ol’ me, the negative one Just wanted to shout out to all the people who feel like they are losing their shit atm.\\ \hline
2020-04-10/29&Such a great idea - not only it would help protect the babies but also making them look so damn cute!\\ \hline
2020-04-10/30&In our latest Marketing Matters video series we uncover some top tips which can help open up new opportunities for your business… In today's FB live (Episode 70), we will discuss the successful model of New Zealand.\\ \hline
2020-04-11/31&Here's all you need to know The Way It's Going Right Now I See It As 50-50!
People are in need!!\\ \hline
2020-04-11/32&Indigenous Groups Isolated by Coronavirus Face Another Threat: Hunger Uncomfortable to say the least, a human rights violation IMO!\\ \hline
2020-04-12/33&We need help with groceries \\& bills.\\ \hline
2020-04-12/34&Steph Curry Mix - "Baby Pluto" Watch our full episode with Dr. John Campbell here: I'm coming across stories like this everywhere.\\ \hline
2020-04-12/35&.maybe just a snippet?...please and thank you ...and please stay safe We need honest products that work for people!!!\\ \hline
2020-04-13/36& “The hard part’s the coming home part.” Earlier this week, Abbott closed all state parks and historic sites and announced new drive-thru testing efforts to battle the COVID-19 crisis.\\ \hline
2020-04-13/37&So again its all about trump!!
Dr. Fauci and President Biden would get all the ratings because people are tired of Trump's shit show.\\ \hline
2020-04-13/38&This thing is seriously infectious Wish more people would listen to this guy.
I can not believe what I’m watching.\\ \hline
2020-04-14/39&It rained all day, \\& I was feeling a little under the weather, so here are pictures of some tchotchke, as well as a few great albums I listened to today.\\ \hline
2020-04-15/40&I think worlds biggest scam is scientists \\& people who discover things which are of no use to human race - like solar system - rather we should have channelise to understand human body better.\\ \hline
2020-04-15/41&The state’s corona count shot up to 730 today An estimated 4,300 women give birth every day in the country.\\ \hline
2020-04-16/42&Mandatory 4 each to stock the Kit. 4S has implemented several protective measures to enable safe delivery of classroom-based training to meet your workplace H\\&S needs.\\ \hline
2020-04-16/43&Congress needs to make sure patients/others have access to health ins, cancer meds at home \\& lower-income patients/survivors can enroll in Medicaid News 12 featured story about COVID-19 plasma therapy trials being offered at Montefiore Nyack Hospital.\\ \hline
2020-04-16/44&(I ask b/c I had a nightmare he nuked all the "uncooperative" states to "slow the spread") "Does Israel Have the Right to Cage Two Million People in a Coronavirus-Ravaged Prison Camp?" Remember when the Government was less inept?\\ \hline
2020-04-17/45&Wait, just yesterday he was saying it’s up to the States when they open, now this...such a fucking idiot and people are going to needlessly die because of his stupidity.\\ \hline
2020-04-17/46&From today, staff from Northants Libraries will be calling vulnerable people to check that they are safe + well during the corona virus pandemic + if they need support with food deliveries, prescriptions or just someone to talk to.\\ \hline
2020-04-18/47&International World Heritage Day. Notwithstanding, let’s all practice social distancing in order to flatten the curve.
The latest The Social Media World!\\ \hline
2020-04-18/48&The quarantine complainers are protesting for the right to spread a deadly virus that has no vaccine or cure and killed 100k people in a few months.\\ \hline
2020-04-18/49&Listen, I think all politicians, countries, governments shouldn't point fingers, it's not the time to blame \\& make excuses.\\ \hline
2020-04-19/50&I need time to grieve People have to find a way to get back to normalcy but with caution.\\ \hline

2020-04-19/51&Estimating COVID-19 Case Fatality  Rates (CFR) Update 9th April:  the CFR is 0.72\% – the lowest end of the current prediction interval and in line with several other estimates.\\ \hline
2020-04-19/52&Beware of fake news, nowadays loads of news are being spread on social media platforms, which are completely baseless and modified.\\ \hline
2020-04-20/53&“In times like these it’s always good to remember there have always been times like these.” It's happening guys.\\ \hline
2020-04-20/54&Tag us and let us know about your co-workers at home:) Speaking to Children about Coronavirus: The new book "Coronavirus - A Book for Children" has been thoughtfully read aloud by Nurse Jane Ferrara.\\ \hline
2020-04-20/55&I just wish I saw more people focusing their time on valuable things like saving the health of the Earth rather than protesting against a decision our government has made to save our lives and the other around us.\\ \hline
2020-04-21/56&If you are a manager who has employees working from home, make sure you reach out to them at least once a day, make them  feel valued during this stressful time.\\ \hline
2020-04-21/57&(1 of 10) As a busy family of 3 we need 2 more donations are available.\\ \hline
2020-04-22/58&He knows what his people needs,he speaks in a language that even the poor can understand.\\ \hline
2020-04-22/59&Check out the video here: The video shows the top COVID19 countries by cases per million people Soonest to reach 1000 in Nigeria ya ALLAH forgive Us, Coronavirus travel upside: More airlines around the world banish the dreaded middle seat.\\ \hline
2020-04-23/60&Unions work hand-in-glove with management A global view of design and urban planning post-COVID-19.\\ \hline
2020-04-23/61&So after a long day of work, I can home to this!\\ \hline
2020-04-24/62&“Please Don't Drink Disinfectant, Lysol And Dettol Maker Said After Trump Suggested People Could Inject It To Kill The Coronavirus.” This is not a test.\\ \hline
2020-04-24/63&I've spoken with many who don't know about the diversion of flights on March 13th We'd just like say thank you for your support during these difficult times!\\ \hline
2020-04-24/64&Life in South Africa will gradually begin to return to normal from next month, with government steadily easing the COVID-19 lockdown regulations, albeit under stringent stipulations.\\ \hline
2020-04-25/65&now its a word I type of say several times an hour “I couldn’t even take steps for probably the first five or six days with two people helping me with a walker.” Happy for this neighbor’s recovery.\\ \hline
2020-04-25/66&I'm sure Hunter would donate big bucks from his partners What it feels like having an idiot run the country.
In good times and bad.\\ \hline
2020-04-26/67&Brazil reports 128 new cases and 12 new deaths bringing total confirmed cases there to 59,324 and 4,057 total deaths.
Totaling to 960,651 cases \\& 54,256 deaths.\\ \hline
2020-04-26/68&Some states, including Maharashtra, will stick to existing curbs for controlling the pandemic Read full e-paper here: People, there is a pandemic going on with lots of people dying.\\ \hline
2020-04-26/69&FCA’s 200-bed field hospital in Brazil is ready to treat COVID-19 patients.
Houston, Texas spent \$17 million on a facility for COVID patients.\\ \hline
2020-04-27/70&American people and the international community need an answer from the US government.\\ \hline
2020-04-27/71&The amount of mental health struggles that have happened as of late, it's constantly growing." Fruits \\& juices for our young COVID-19 patients at files hospital.\\ \hline
2020-04-27/72&Good Morning Happy Monday Yes another new week Brighter today, definitely a two steps forward one back recovery.\\ \hline
2020-04-28/73&(By the way, that's how real sarcasm works.) World Needs To Know This.
Who knew World War III would look like this?\\ \hline
2020-04-28/74&When is the right time to open the economy back up?\\ \hline
2020-04-28/75&Here are the ways they see the pandemic transforming our societies: Here’s a short song about the impacts of the Corona virus.
That's what viruses do.\\ \hline
2020-04-29/76&WestSidewalks (BchSide) Closed-Off. We have considerations for auditors and audit committees to ensure continued high quality financial reporting: Thanks to RZA and the Children’s Literacy Society… I'm Canadian, but I like Gov. Kristi Noem's plan!\\ \hline
2020-04-29/77&It’s like one continuous day right?
I got an idea Rest in Peace Signed Coronavirus, what coronavirus?
Day 46.\\ \hline
2020-04-29/78&I think care homes get money from the Government for people dying of COVID 19 so they have an incentive.\\ \hline
2020-04-30/79&This is what good government looks like.
This is what good planning and information looks like.\\ \hline
2020-04-30/80&Thank You. Morning guys, hope you're having a great week so far.\\ \hline

    \caption{Summarized topics for each day; The top largest clusters has been selected for summarizing.}
    \label{tab:summarized_topics_full}

\end{longtable}

\end{document}